\documentclass[final]{cvpr}

\usepackage{times}
\usepackage{epsfig}
\usepackage{graphicx}
\usepackage{amsmath}
\usepackage{amssymb}
\usepackage{booktabs}
\usepackage{enumitem}
\usepackage{multirow}
\usepackage[belowskip=-5pt,aboveskip=2pt]{caption}

\usepackage[pagebackref=true,breaklinks=true,colorlinks,bookmarks=false]{hyperref}

\def\confYear{CVPR 2021}
\newcommand{\nbr}{SVR}

\newcommand{\ir}{IVR}

\newcommand{\epic}{EPIC-KITCHENS}
\newcommand{\msr}{MSR-VTT}
\newcommand{\yc}{YouCook2}

\newcommand{\myparagraph}[1]{\vspace{0.1cm}\noindent\textbf{#1}}
\newcommand{\exquote}[1]{\textit{``#1''}}

\begin{document}

\title{On Semantic Similarity in Video Retrieval}

\author{Michael Wray \qquad \quad
Hazel Doughty\thanks{Now at University of Amsterdam.}
\qquad \quad
Dima Damen\\
Department of Computer Science, University of Bristol, UK
}

\maketitle

\begin{abstract}
\vspace{-0.7em}
Current video retrieval efforts all found their evaluation on an instance-based assumption, that only a single caption is relevant to a query video and vice versa. 
We demonstrate that this assumption results in performance comparisons often not indicative of models' retrieval capabilities.
We propose a move to semantic similarity video retrieval, where (i)~multiple videos/captions can be deemed equally relevant, and their relative ranking does not affect a method's reported performance and (ii) retrieved videos/captions are ranked by their similarity to a query. We propose several proxies to estimate semantic similarities in large-scale retrieval datasets, without additional annotations. Our analysis is performed on three commonly used video retrieval datasets (MSR-VTT, YouCook2 and EPIC-KITCHENS).
\end{abstract}
\vspace{-1em}

\section{Introduction}

Video understanding approaches which incorporate language have demonstrated success in multiple tasks including  captioning~\cite{venugopalan2015sequence,yao2015describing}, video question answering~\cite{zeng2016leveraging,zhu2017uncovering} and navigation~\cite{anderson2018vision, gupta2017cognitive}.
Using language to search for videos has also become a popular research problem, known as video retrieval. Methods learn an underlying multi-modal embedding space to relate videos and captions.
Along with large-scale datasets~\cite{damen2020rescaling,miech2019howto100m,rohrbach2015dataset,xu2016msr-vtt,zhou2017towards}, several video retrieval benchmarks and challenges~\cite{albanie2020bench,zhou2020bench} compare state-of-the-art, as methods inch to improve evaluation metrics such as Recall@K and Median Rank.

In this paper, we question the base assumption in all these datasets and benchmarks---that the only relevant video to a caption is the one collected with that video.
We offer the first critical analysis of this assumption, proposing semantic similarity relevance, for both evaluation and training. 
Our effort is inspired by works that question assumptions and biases in other research problems such as VQA~\cite{goyal2017making,zhang2016yin}, metric learning~\cite{musgrave2020metric}, moment retrieval~\cite{otani2020challengesmr}, action localisation~\cite{alwassel2018diagnosing} and action recognition~\cite{choi2019can,li2019repair,moltisanti2017trespass}.

\begin{figure}[t]
    \centering
    \includegraphics[width=1\linewidth]{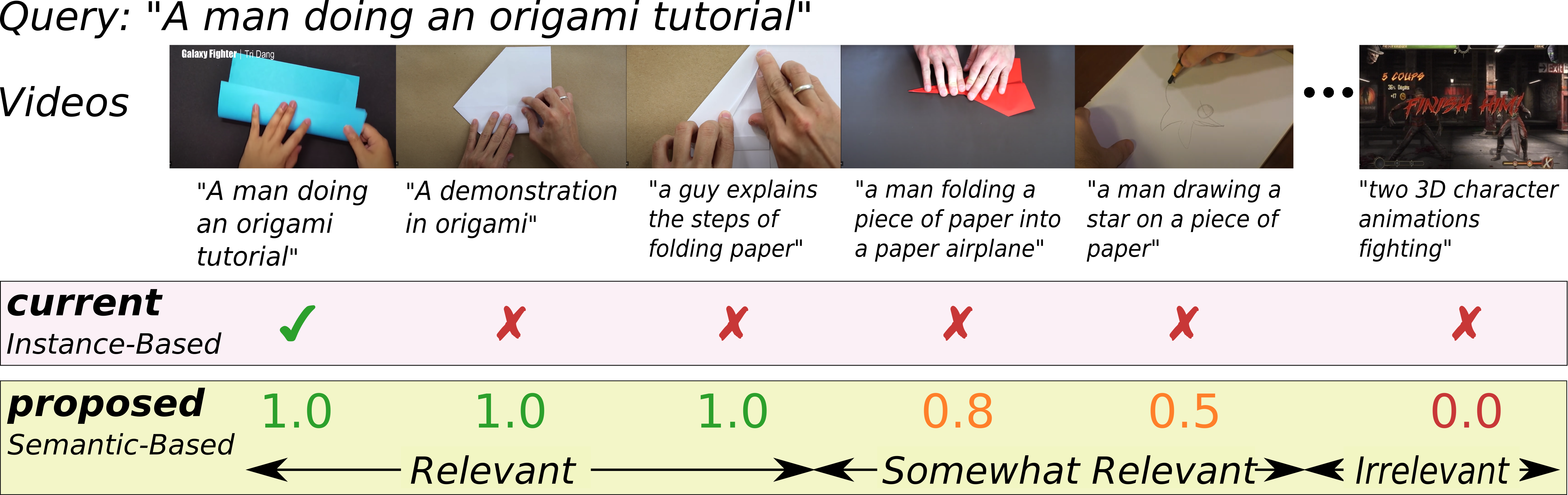}
    \caption{All current video retrieval works treat caption collected for a certain video as relevant, even when other videos are equally relevant to a query text. This makes the evaluation of popular datasets ad hoc at times. We propose to use continuous similarity, allowing multiple videos to be treated as equally relevant. 
    Ex. from MSR-VTT~\cite{xu2016msr-vtt}.\vspace{-4pt}}
    \label{fig:intro-fig}
\end{figure}

As shown in Fig.~\ref{fig:intro-fig}, current approaches target instance-based retrieval---that is, given a
query caption such as \exquote{A man doing an origami tutorial}, only one origami video is considered as the correct video to retrieve. In fact, many videos within the dataset can be similar to the point of being identical.
The order in which such videos are retrieved should not affect the evaluation of a method.
Instead, we propose utilising semantic similarity between videos and captions, where we
assign a similarity score between items of differing modalities.
This allows multiple videos to be considered relevant to a caption and provides a way of ranking videos from most to least similar.

Our contributions can be summarised:
(i) We expose the shortcoming of instance retrieval in current video retrieval benchmarks and evaluation protocols.
(ii) We propose video retrieval with semantic similarity, both for evaluation and training, where videos are ranked by their similarity to a caption, allowing multiple videos to be considered relevant and vice-versa.
(iii) Avoiding large annotation effort, we propose several proxies to predict semantic similarities, using caption-to-caption matching.
(iv) We analyse three benchmark datasets, using our semantic similarity proxies, noting their impact on current baselines and evaluations.

\section{Related Work}
\vspace{-3pt}
\noindent We review image retrieval works that use semantic knowledge then discuss current approaches to video retrieval.
\vspace{-3pt}

\subsection{Semantic Image Retrieval} 
\vspace{-6pt}

While most works focus on instance-based retrieval, a few works have explored semantic-based image retrieval. 

Early works attempted to manually annotate small-scale datasets with semantic knowledge.
Oliva~\textit{et al.}~\cite{oliva1999global} defined three axes of semantic relevance (e.g. artificial vs natural) 
in order to relate images.
Using categories instead, Ojala~\textit{et al.}~\cite{ojala2001semantic} asked annotators to split images within a dataset into discrete categories.
They then considered all images within the same category as relevant.

In their investigative work, Enser~\textit{et al.}~\cite{enser2007facing} showcase that semantic relevance cannot be gleaned from images alone, as it requires the knowledge of places, societal class~\textit{etc.}
Barz and Denzler~\cite{barz2019hierarchy} draw a similar conclusion that visual similarity does not imply semantic similarity and so project images into a class embedding space learned from WordNet~\cite{miller1995wordnet}.
Chen~\textit{et al.}~\cite{chen2019cross} instead learn two spaces, one for images and one for text, with the notion that features in either space should be consistent if they are semantically relevant.
Gordo and Larlus~\cite{gordo2017beyond} train their model for image-to-image retrieval with the notion of semantic relevance.
By learning an embedding using semantic proxies (METEOR~\cite{banerjee2005meteor}, tf-idf and Word2Vec~\cite{mikolov2013efficient}) defined between image captions, they show that semantic knowledge improves retrieval performance.
Concurrent with out work, Chun~\textit{et al.}~\cite{chun2021probabilistic} highlight the issue of instance based evaluation for cross-modal retrieval in images.
They propose using R-Precision as an evaluation metric incorporating further plausible matches via class knowledge.
However, all these works still use binary relevance for training and evaluation, \ie an image/caption is either relevant or not, excluding images which may be somewhat relevant.

Closest to our work, Kim~\textit{et al.}~\cite{kim2019deep} explore non-binary relevance in image retrieval.
They propose a log-ratio loss in order to learn a metric embedding space without requiring binary relevance between items.
Their work is primarily focused on human pose, in which they use the distance between joints to rank images.
They also explore within-modal image retrieval using word mover's distance, as a proxy 
for semantic similarity.
Up to our knowledge, \cite{kim2019deep} offers the only prior work, albeit in image retrieval, 
to investigate both training and evaluating relevance which extends beyond both binary and instance-based relevance.

\subsection{Video Retrieval}

Early works in video retrieval simply extended image-retrieval approaches by temporally aggregating frames for each video~\cite{dong2018predicting,otani2016learning,torabi2016learning,xu2015jointly}.
These works are attributed for defining the cross-modal video retrieval problem and standard evaluation metrics.
In qualitative results, they argue models are superior if they retrieve multiple relevant videos, despite the quantitative metrics only evaluating the corresponding video.

With larger datasets becoming available~\cite{anne2017localizing,krishna2017dense,miech2019howto100m,oncescu20queryd,wang2019vatex,xu2016msr-vtt,zhou2017towards},
methods focused on using self-supervision~\cite{alayrac2020self,rouditchenko2020avlnet,zhu2020actbert}, sentence disambiguation~\cite{chen2019cross,wray2019fine}, multi-level encodings~\cite{dong2020hybrid,yu2018joint}, mixing \exquote{expert} features from pre-trained models~\cite{gabeur2020multi,liu2019use,miech2018learning,mithun2018learning} and weakly-supervised learning from massive datasets~\cite{miech2020end, miech2019howto100m,patrick2020support}.
All these works train and evaluate for instance-based video retrieval.

Two recent works explored using semantic similarity during training~\cite{patrick2020support,wray2019fine}.
Our previous work~\cite{wray2019fine} uses class knowledge to cluster captions into relevance sets for triplet sampling.
Patrick~\textit{et al.}~\cite{patrick2020support} propose a captioning loss, where the embedding caption is re-constructed from a support set of videos. This ensures shared semantics are learned between different instances and gives large improvements when the support set does not include the corresponding video---forcing the model to generalise. However, this work is evaluated using instance-based retrieval. 

This paper is the first to 
scrutinise current benchmarks in video retrieval, which assume instance-based correspondence. We propose semantic similarity video retrieval as an alternative task, for both evaluation and training.

\begin{table*}[t]
    \centering
    \resizebox{\linewidth}{!}{
    \begin{tabular}{lcrrccccc}
        \toprule
         & Text Type & \# Captions & Test Set Size $\downarrow$ & Scenario & Source & Eval. Metrics & Semantic Info  \\
         \midrule
         MSVD~\cite{chen2011collecting} & Caption & *86k & 670 & Open & {Y}ou{T}ube & Recall@k, Avg. Rank & Multi-Lang. \\
         MPII movie~\cite{rohrbach2015dataset} & Script & 64k & 1,000& Movie Scripts & Movies & Recall@k, Avg. Rank & None \\
         DiDeMo~\cite{anne2017localizing} & Caption & 40k & 1,004 & Open & Flickr & Recall@k & None \\
         \msr~\cite{xu2016msr-vtt} & Caption & 200k & 2,990 & Open & {Y}ou{T}ube & Recall@k, Avg. Rank & Category \\
         \yc~\cite{zhou2017towards} & Caption & 15k & 3,310 & Cooking & YouTube & Recall@k, Avg. Rank & None \\
         QuerYD~\cite{oncescu20queryd} & Caption & 31k & 4,717 & Open & YouTube & Recall@k, Avg. Rank & Category \\
         ActivityNet+Captions~\cite{krishna2017dense} & Dense Captioning & 100k & 4,917 & Daily Living & YouTube & Recall@k & None \\ 
         TVR~\cite{lei2020tvr} & Video Subtitle & 109k & 5,445 & TV Shows & TV & Recall@k & TV Show \\
         Condensed Movies~\cite{bain2020condensed} & Caption & 34k & 6,581 & Movies & YouTube & Recall@k, Avg. Rank & Movie \\
         VATEX~\cite{wang2019vatex} & Caption & *412k & 8,920 & Open & YouTube & Recall@k, Avg. Rank & Multi-Lang. \\
         \epic~\cite{damen2020rescaling} & Short Caption & 77k & 9,668 & Kitchen & Egocentric & mAP,nDCG & Action Class  \\
         
         \bottomrule
    \end{tabular}
    }
    \caption{Details of popular Datasets in Video Retrieval, ordered by test set size. *Number of English captions. }
    
    \label{tab:retrieval_datasets}
\end{table*}
\begin{figure*}[t]
    \centering
    \includegraphics[width=1\textwidth]{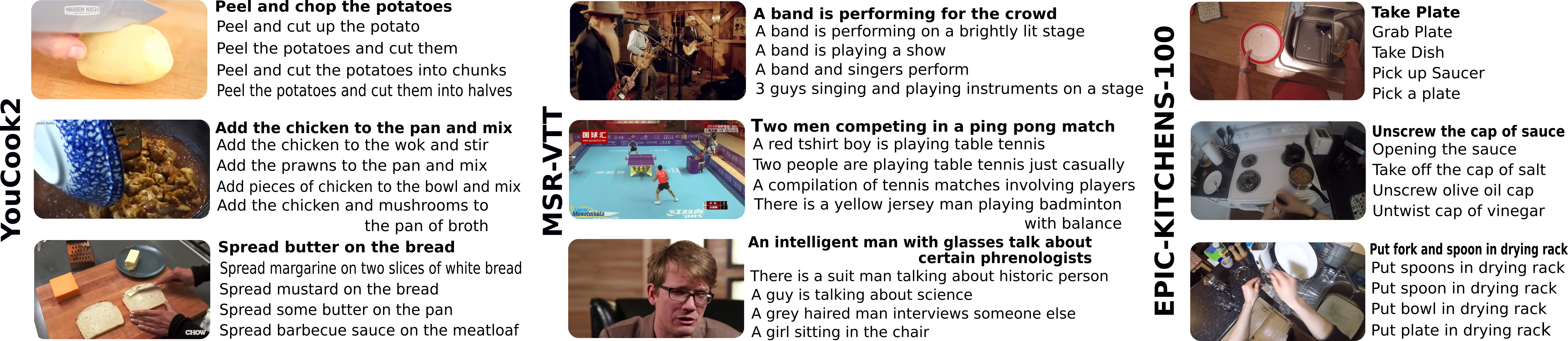}
    \caption{Video examples from the test set of three datasets showcasing the corresponding caption (bold) used as ground-truth along with highly relevant captions for other videos in the test set, considered irrelevant by \ir{}. In fact at times, such as the top example from MSR-VTT, a caption deemed irrelevant by the benchmarks can be a more specific description of the video.}
    \label{fig:qual_caption_issues}
\end{figure*}

\section{Shortcomings of Current Video Retrieval Benchmarks}
\label{sec:instance_analysis}

In this section, we formalise the current approaches to video retrieval, and highlight the issues present with their Instance-based Video Retrieval (\ir{}) assumption, which impacts the evaluation of common benchmark datasets. 

Formally, given a set of videos $X$ and a corresponding set of captions $Y$, current approaches define the 
similarity $S_I$ between a video $x_i$ and a caption $y_j$ which captures this one-to-one relationship. For each video/caption there is exactly one relevant caption/video:
\vspace{-8pt}
\begin{equation}
\vspace{-6pt}
    S_I(x_i, y_j) = 
        \begin{cases}
            1, & \text{if}\ i == j\\
            0, & \text{otherwise}
        \end{cases}
    \label{eq:ibr_similarity}
\end{equation}
Alternatively, if multiple captions are collected per video as in~\cite{xu2016msr-vtt}, then we consider the caption $y_{j,k}$ as the $k^{th}$ caption of the $j^{th}$ video.
As in Eq.~\ref{eq:ibr_similarity}, this only considers captions of the corresponding video to be relevant.

\ir{} relies on the correspondence between the video and the caption captured during dataset collection.
This is typically a caption provided by an annotator or transcribed from the video's narration.
Importantly, the above formulation makes the assumption that no two captions of different videos are relevant-enough to impact the evaluation or training of retrieval methods. 
We start by qualitatively examining this assumption for current benchmarks.

\myparagraph{Datasets } In Table~\ref{tab:retrieval_datasets} we show the statistics of datasets that are actively being used as benchmarks for video retrieval.
We order these by the size of the test set, as a larger test set is not only challenging in distinguishing between more examples, but importantly increases the chance of having other relevant items, beside the corresponding video/caption. 

Most datasets~\cite{chen2011collecting,krishna2017dense,xu2016msr-vtt,zhou2017towards} have been collected from {Y}ou{T}ube and annotated after the fact via crowd-sourcing.
Notably, MPII movie~\cite{rohrbach2015dataset} instead used movie scripts as captions for each of the video clips and \epic{} utilised transcribed audio narrations provided by the video collectors.
However, in all cases, the captions were collected  with the annotator observing a single video, thus a caption's relevance to other videos could not be considered. 

\msr~\cite{xu2016msr-vtt}, MSVD~\cite{chen2011collecting} and VATEX~\cite{wang2019vatex} include multiple captions per video, from multiple annotators, due to the datasets being collected for captioning and paraphrase evaluation. 
Nevertheless, during evaluation, prior works~\cite{miech2018learning,mithun2018learning,yu2018joint} all use a test set that considers only one caption per video. While some works~\cite{chen2020fine,gabeur2020multi,liu2019use,miech2018learning,wray2019fine} utilise multiple captions during training, captions are only relevant to the corresponding video and considered irrelevant to all other videos. 
The video pentathlon~\cite{albanie2020end} recently defined a retrieval challenge across five datasets~\cite{anne2017localizing,chen2011collecting,krishna2017dense,xu2016msr-vtt,zhou2017towards}.
This pentathlon similarly utilises \ir.

We focus our analysis on three datasets with a large test set, \msr{}, \yc{} and \epic{}. We consider \yc{} and \epic{} as these focus on the single scenario of cooking, increasing the number of relevant captions within the dataset.

\myparagraph{Qualitative Analysis } 
We start by highlighting stark qualitative examples, showcasing the shortcomings of the \ir{} assumption, in Fig.~\ref{fig:qual_caption_issues}. 
For each video, we show a key frame along with five captions from the test set. We highlight the corresponding caption according to the dataset annotations in bold---which is used as ground-truth for evaluating and ranking various methods.
In each case, we show several indistinguishable captions that are all relevant in describing the corresponding video.
In fact, identifying which caption is the ground truth would be challenging for a human. However, a method that \textit{potentially randomly} gets the bold captions higher in the retrieval list would be considered state-of-the-art, while another might be unfairly penalised. 
These valid captions contain synonyms, a change in the sentence structure or more/less details in describing the video.

Additionally, we find captions which are not  interchangeable but are still somewhat relevant to the video.
For instance, the second example of \epic{} includes captions of opening other bottles---\eg sauce bottle vs the vinegar/oil bottle. 
These captions
should be ranked higher than an irrelevant caption (\eg ``cutting a tomato'').

\myparagraph{Conclusion } While the concern with \ir{} is clarified in this section, the task of manually annotating all relevant captions, as well as somewhat relevant captions, is unachievable due to time and cost required.
Instead, we propose several proxy measures for semantic similarity between videos and captions, which require no extra annotation effort and 
 use external corpora or knowledge bases.

\section{Video Retrieval with Semantic Similarity}
\label{sec:semantic-similarity}

In this paper, we propose to move beyond Instance-based Video Retrieval (\ir{}) towards video retrieval that uses semantic similarity between videos and captions, for both video-to-text and text-to-video retrieval.
We first define Semantic Similarity Video Retrieval~(\nbr), then propose an evaluation protocol, as well as an approach to incorporate semantic similarity during training. 
Finally, in Sec.~\ref{subsec:sem_sim_proxy_def} we propose multiple approaches to estimate semantic similarity from captions without the need for manual annotations.

\subsection{Definition}
\label{subsec:sem_sim_def}

Given the set of videos, $X$, and a corresponding set of captions, $Y$.
We define a semantic similarity function, $S_S(x_i, y_j) \rightarrow [0, 1]$, which calculates a continuous score that captures the 
similarity between any (video, caption) pair.
Similar to \ir, $S_S(x_i, y_j) = 0$ if the caption is irrelevant to the video and $1$ for maximally relevant.
Different from \ir{}, multiple captions can have a similarity of $1$ to a video, and analogously for videos.
Additionally, the continuous value of $S_S$ can model varying levels of similarity.
If $S_S(x_i, y_j) > S_S(x_i, y_k)$ then $y_j$ is a more relevant caption to the video $x_i$ than the caption $y_k$. Consequently, if $S_S(x_i, y_j) = S_S(x_i, y_k)$ then both captions are considered equally relevant and retrieving them in any order should not be penalised by the evaluation metric.

\subsection{Evaluation}
\label{subsec:sem_sim_eval}

To accommodate for cross-modal retrieval, i.e. both text-to-video and video-to-text, we use the terms ``item'' and ``query'' to refer to either a video or a caption.
For a given query, all items from the opposing modality are ranked according to their distance from the query in the learnt embedding space.
Benchmarks in \ir{} use the following evaluation metrics: Recall@K, Geometric Mean\footnote{Geometric Mean averages Recall@K over a range, typically $\{1, 5, 10\}$, each giving the percentage of queries for which the corresponding item was found within the top K results.} and Average Rank (median or mean) of the corresponding item. 

In \nbr{}, a different evaluation metric is needed due to limitations of all current evaluation metrics used for \ir.
Firstly, Average Rank only allows for a single relevant item. 
Whilst Recall@K can be used to evaluate queries with multiple items, a threshold on the continuous similarity is required. Additionally, choosing the value of K has to be considered carefully.
If the value of K is less than the number of relevant items for a given query, the metric would not be suitable to assess a model's true performance.
This is a concern for \nbr{} where the number of relevant items will vary per query, resulting in an unbalanced contribution of different query items to the metric.
Mean Average Precision (mAP) has also been used for retrieval baselines as it allows for the full ranking to be evaluated. However, mAP also requires binary relevance between query and items.

We seek an evaluation metric which is able to capture multiple relevant items and take into account relative non-binary similarity.
We thus propose using normalised Discounted Cumulative Gain (nDCG)~\cite{jarvelin2002cumulated}.
nDCG has been used previously for information retrieval~\cite{chapelle2010gradient, radlinski2010comparing}. It requires similarity scores between all items in the test set.
We calculate Discounted Cumulative Gain (DCG) for a query $q_i$ and the set of items $Z$, ranked according to their distance from $q_i$ in the learned embedding space:
\vspace{-5pt}
\begin{equation}
\vspace{-8pt}
    DCG(q_i) = \sum_{j=1}^{|\mathcal{R}_{q_i}|} \frac{2^{S_S(q_i, z_j)} - 1}{log(j+1)}
\end{equation}
where $\mathcal{R}_{q_i}$ is the set of all items of the opposing modality, excluding irrelevant items, for query ${q_i : \mathcal{R}_{q_i} = \{z_j |  S_S(q_i, z_j) > 0,  \forall z_j \in Z\}}$\footnote{Note that nDCG does not penalise the case when a large number of low-relevant items are present. This can be alleviated by thresholding $S$.}.
Note that this equation would give the same value when items of the same similarity $S_S$ are retrieved in any order.
It also captures differing levels of semantic similarity.

nDCG can then be calculated by normalising the DCG score such that it lies in the range $[0, 1]$: ${nDCG(q_i) = \frac{DCG(q_i)}{IDCG(q_i)}}$
where $IDCG(q_i)$ is calculated from $DCG$ and $Z$ ordered by relevance to the 
query $q_i$.

For overall evaluation, we consider both video-to-text and text-to-video retrieval and evaluate a model's nDCG as:
\vspace{-4pt}
\begin{equation}
\resizebox{\linewidth}{!}{
    $nDCG(X, Y) = \frac{1}{2|X|} \sum\limits_{x_i \in X} nDCG (x_i) +  \frac{1}{2|Y|} \sum\limits_{y_i \in Y} nDCG (y_i)$}
    \label{eq:fullNDCG}
\end{equation}
Note that Eq.~\ref{eq:fullNDCG} allows for a different number of videos and captions in the test set.

\subsection{Training}
\label{subsec:sem_sim_training}
\vspace*{-6pt}

In addition to utilising semantic similarity for evaluation, it can also be incorporated during training. 
A contrastive objective can be defined to learn a multi-modal embedding space, e.g. the triplet loss: 
\vspace{-2pt}
\begin{equation}
\resizebox{\linewidth}{!}{
  $ L_{t}(x_i, y_j, y_k) = \max\left(m + D(f({x_i}), g({y_j})) - D(f({x_i}), g({y_k})), 0\right)$}
    \label{eq:triplet_loss}
\end{equation}
where $D(\cdot, \cdot)$ is a distance function, $f(\cdot)$ and $g(\cdot)$ are embedding functions for video and text respectively, and $m$ is a constant margin.
In \ir{}, the triplets $x_i$, $y_j$ and $y_k$ are sampled such that $S_I(x_i, y_j)=1$ and $S_{I}(x_i, y_k)=0$ (see Eq.~\ref{eq:ibr_similarity}).
In \nbr{}, we use triplets such that $S_S(x_i, y_j)\geq T$ and $S_S(x_i, y_k)<T$ where $T$ is a chosen threshold.

\myparagraph{Alternative Losses } 
Other alternatives to the triplet loss can be utilised, such as the approximate nDCG loss from~\cite{qin2010general}, log-ratio loss from~\cite{kim2019deep}, or losses approximating mAP~\cite{brown2020smoothap,cakir2019deep,qin2010general,revaud2019learning}. 
It is worth noting that some of these works combine the proposed loss with the instance-based triplet loss for best performance~\cite{brown2020smoothap,kim2019deep}. Additionally, approximating mAP requires thresholding as mAP expects binary relevance.
Note that all these works, apart from~\cite{kim2019deep}, attempt instance-based image retrieval. Experimentally, we found the log-ratio loss to produce inferior results to thresholding the triplet loss.
Adapting these losses to the \nbr{} task is an exciting area for exploration in future work. 

\subsection{Proxy Measures for Semantic Similarity}
\label{subsec:sem_sim_proxy_def}
Collecting semantic similarity from human annotations, 
for all but the smallest datasets, is costly, time consuming\footnote{Annotators would have to observe a video with two captions and indicate their relative relevance. For $n$ videos and $m$ captions this is $O(nm^2)$.} and potentially noisy.
Previous work in image retrieval~\cite{gordo2017beyond} demonstrated that semantic similarities of captions can be successfully utilised. 
We use the knowledge that each video in the dataset was captured with a corresponding caption, which offers a suitable description of the video, and thus use the semantic similarity between captions instead, 
\ie we define $S_S(x_i, y_j)$ as
\vspace{-8pt}
\begin{equation}
\vspace{-4pt}
    S_S(x_i, y_j) =
    \begin{cases}
    1 & i == j\\
    S'(y_i, y_j) &otherwise
    \end{cases}
    \label{eq:nbr}
\end{equation}
where $S'$ is a semantic proxy function relating two captions.

We define four semantic similarity measures which we use to compute $S'(y_i, y_j)$---based on bag of words, part-of-speech knowledge, synset similarity and the METEOR metric~\cite{banerjee2005meteor}. 
We choose these proxy measures such that they should scale with the size of the dataset, not requiring any extra annotation effort, but acknowledge that some datasets may be better suited by one proxy over others.
We investigate this qualitatively and quantitatively in Sec.~\ref{subsec:proxy_measures_comp}.

\myparagraph{Bag-of-Words Semantic Similarity }
Naively, one could consider the semantic similarity between captions as the overlap of words between them.
Accordingly, we define the Bag-of-Words (BoW) similarity as the Intersection-over-Union (IoU) between sets of words in each caption:
\vspace{-8pt}
\begin{equation}
\vspace{-6pt}
    S'_{BoW}(y_i, y_j) = \frac{|w_i \cap w_j|}{|w_i \cup w_j|}
\end{equation}
where $w_i$ and $w_j$ represent the sets of words, excluding stop words, corresponding to captions
$y_i$ and $y_j$.

This proxy is easy to calculate, 
however, as direct word matching is used with no word context. This raises two issues: firstly, synonyms for words are considered as irrelevant as antonyms, \ie \exquote{put} and \exquote{place}.
Secondly, words are treated equally---regardless of their part-of-speech, role in the caption, or how common they are.
Word commonality is partially resolved by removing stop words\footnote{We find using tf-idf to remove/re-weight words comparable to removing stop words in the analysed datasets.}. 
We address the other concerns next. 

\myparagraph{Part-of-Speech Semantic Similarity }
Verbs and nouns, as well as adjectives and adverbs, describe different aspects of the video and as such words can be matched within their part-of-speech.
Matching words irrespective of part-of-speech can result in incorrect semantic similarities.
For example, the captions ``watch a play'' and ``play a board game''.
Alternatively, adverbs can be useful to determine how-to similarities between captions~\cite{doughty2020action}.
By augmenting the part-of-speech, we can ensure that the actions and objects between two videos are similar.

To calculate the Part-of-Speech (PoS) word matching, captions are parsed, and we calculate the IoU between the sets of words for each of the parts-of-speech and average over all parts-of-speech considered.
\vspace{-6pt}
\begin{equation}
\vspace{-6pt}
    S'_{PoS}(y_i, y_j) = \sum_{p \in P} \alpha^p \frac{|w_i^p \cap w_j^p|}{|w_i^p \cup w_j^p|}
\end{equation}
where $p$ is a part-of-speech from the set $P$, $w_i^p$ is the set of words from caption $y_i$ which have a part-of-speech $p$, and $\alpha^p$ is the weight assigned to $p$ such that $\sum_{p \in P} \alpha^p = 1$. 

\myparagraph{Synset-Aware Semantic Similarity}
So far, the proxies above do not account for synonyms, \eg{} \exquote{put} and \exquote{place}, \exquote{hob} and \exquote{cooker}.
We extend the part-of-speech similarity detailed above using semantic relationship information from synsets, i.e. grouped synonyms, from WordNet~\cite{miller1995wordnet} or other semantic knowledge bases.
We modify the part-of-speech proxy, 
\vspace{-6pt}
\begin{equation}
\vspace{-6pt}
    S'_{SYN}(y_i, y_j) = \sum_{p \in P} \alpha^p \frac{|C_i^p \cap C_j^p|}{|C_i^p \cup C_j^p|}
\end{equation}
where $C_i^p$ is the set of synsets within the part-of-speech $p$ for caption $y_i$.
Note that $|C_i^p| \leq |w_i^p|$ as multiple words are assigned to the same synset due to the similar meanings.

\myparagraph{METEOR Similarity }
The first three similarity functions break the sentence into its individual words, with/without parsing knowledge.
Instead, captioning works have proposed metrics that preserve the structure of the sentence, comparing two captions accordingly.
Multiple metrics have been proposed (e.g. BLEU~\cite{papineni2002bleu}, ROUGE~\cite{lin2004rouge} or CIDEr~\cite{vedantam2015cider}) including METEOR~\cite{banerjee2005meteor}.
Originally used for machine translation and later image captioning, Gordo and Larlus~\cite{gordo2017beyond} proposed METEOR as one of their proxy measures for relating images via their captions.

METEOR calculates similarity both via matching, using synsets to take into account synonyms, and via sentence structure, by ensuring that matched words appear in a similar order.
The proxy is then defined as: $S'_{MET}(y_i, y_j) = M(y_i, y_j)$, where $M(\cdot, \cdot)$ is the METEOR scoring function.

\myparagraph{Other proxies}
Other similarity measures, including the use of word/sentence embedding models such as BERT~\cite{devlin2018bert}, do not provide useful similarity scores on video retrieval datasets. This is further discussed in supplementary. 

\section{Semantic Similarity Analysis}
\label{sec:experiments}

\begin{figure*}[t]
    \centering
    \includegraphics[width=.95\linewidth]{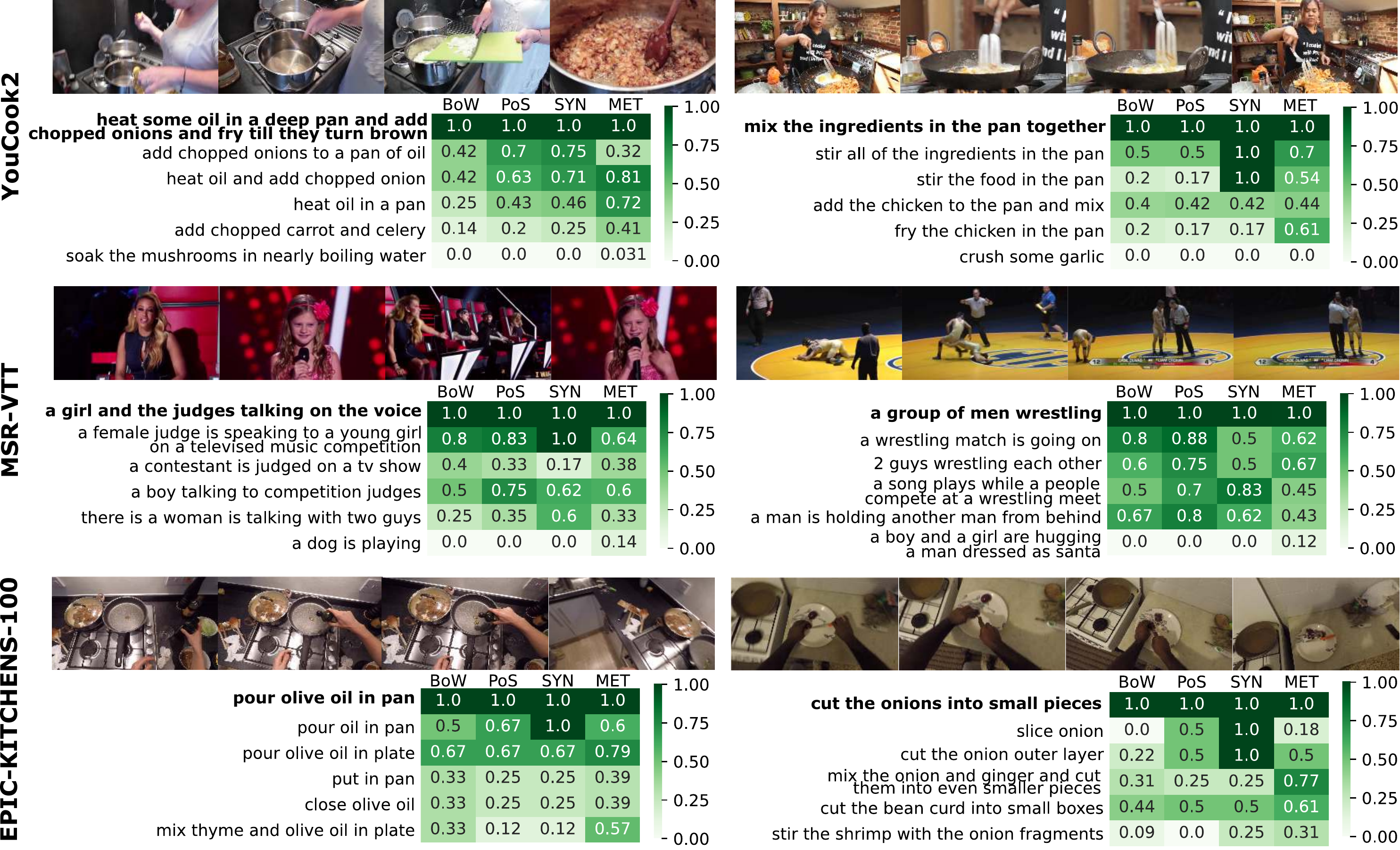}
    \caption{Examples of the proposed semantic similarity proxies (Sec~\ref{subsec:sem_sim_proxy_def}). Captions are shown alongside the score $S'_S(y_i, y_j)$ when compared to the corresponding caption (bold). While scores differ, methods agree on highly-(ir)relevant captions.\vspace*{-5pt}}
    \label{fig:sem_sim_qual_examples}
\end{figure*}

We evaluate baseline methods on the three datasets, with the aim of answering the following questions:
(i) How do the different proxy measures compare to each other on the three datasets?
(ii) What is the impact of the noted shortcomings of \ir{} on methods' performance?
(iii) How do current methods perform when using \nbr{} evaluation for the four proposed proxy measures?
(iv) How does training the models for \nbr{} affect the results?

Next, we present information on the datasets and baseline methods along with their implementation details.

\myparagraph{Datasets} We continue exploring the three public datasets from Sec.~\ref{sec:instance_analysis}. 
These are: 
frequently-used \msr~\cite{xu2016msr-vtt} and \yc~\cite{zhou2017towards}, as well as recently released \epic-100~\cite{damen2020rescaling}.
The latter also has the benefit in that it offers semantic annotations as we show next. 

\myparagraph{Baselines } 
We train a simple embedding baseline with a multi-layer perceptron for each modality which we name as Multi-Modal Embedding or \textbf{MME}.
We additionally consider three publicly available baselines for the aforementioned datasets. 
We use the benchmark implementations provided by the video pentathlon challenge~\cite{albanie2020end} for \msr{} and \yc: \textbf{MoEE~\cite{miech2018learning}:} Multiple video features are extracted from `video experts' and an embedding is learned for each. The final embedding is learned as a weighted sum, determined by the caption. 
\textbf{CE~\cite{liu2019use}:}~Video expert features are passed through a paired collaborative gating mechanism before the embedding and resulting weighted sum.
For \epic{}, we use the baseline method
\textbf{JPoSE~\cite{wray2019fine}:} this trains separate embedding spaces for each part-of-speech in the caption before being combined into a retrieval embedding space.
Implementation details match the publicly available code per baseline as trained for \ir.

\myparagraph{Parsing and Semantic Knowledge }
We parse the captions using Spacy's large web model~\cite{spacy}. We limit these  to verbs and nouns, setting $\alpha^p = 0.5$ for each in all experiments.
When computing the Synset-Aware Similarity, we use the synsets released as part of~\cite{damen2020rescaling} for both \epic{} and \yc{}, as both share the domain of cooking. We found that the synset information transfers well across both datasets. 
Synset knowledge for MSR-VTT is found using WordNet~\cite{miller1995wordnet} and the Lesk algorithm~\cite{lesk1986automatic}.
\msr{} includes multiple captions per video, therefore, for robust word sets, we only include words which are present in 25\% or more of all of the captions for a given video (excluding stop words).
For METEOR, we use the NLTK implementation~\cite{loper2002nltk}. Additionally, to calculate $S_{MET}$ for \msr, we use many-to-many matching with a non-Mercer match kernel~\cite{lyu2005mercer}.

\begin{figure*}[t]
    \centering
    \includegraphics[width=.95\linewidth]{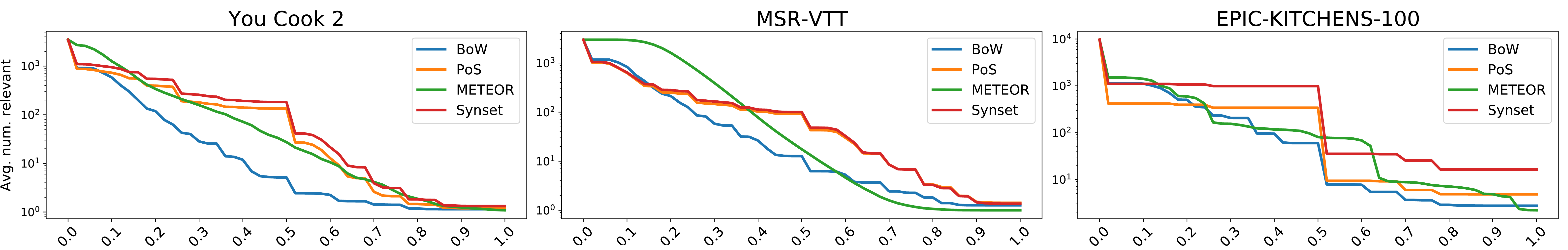}
    \caption{Average number of relevant captions for a video with a given threshold over each dataset and proxy measure.}
    \label{fig:sem_histogram}
\end{figure*}

\vspace{-2pt}
\subsection{Proxy Measure Comparisons}
\label{subsec:proxy_measures_comp}

We first clarify differences between the  semantic similarity proxies with qualitative examples. 
Fig.~\ref{fig:sem_sim_qual_examples} shows examples from \yc{}, \epic{} and \msr{}.

BoW is the tightest proxy to \ir, only considering captions as equally relevant when the set of words match exactly.
The Synset proxy is the only one to consider
the captions ``stir food in the pan'' and ``mix the ingredients in the pan together'' equivalent. This is because it separately focuses on the verb and noun (similar to PoS) and is able to relate words such as \exquote{stir} and \exquote{mix}. While METEOR also considers synonyms, it aims for a good alignment in word order, therefore it gives all captions containing \exquote{in the pan} a high score, even when the verb differs. This also explains the low score given to ``add chopped onions to a pan of oil'' compared to PoS and SYN even though the caption contains many of the same concepts.

For \msr, we show examples that demonstrate limitations of semantic proxies. All proxies rank the caption \exquote{a boy talking...} higher than \exquote{a contestant is judged...}. Similarly, the relevance of the caption \exquote{a song plays} requires access to the audio associated with the video and cannot be predicted from the associated caption.

Having established an understanding of the proxies, we now quantitatively assess them.
In Fig.~\ref{fig:sem_histogram}, 
we calculate the similarity between a video and all captions in the dataset using Eq.~\ref{eq:nbr}.
We then vary the threshold, $T$, for each proxy and compute the number of captions where $S(x_i, y_j) \ge T$.
We plot the average number of `relevant' captions as the threshold increases.
Note the y-axis is a log scale.
In all cases, we note that even at high thresholds the average number of relevant captions is higher than 1.
As expected, the synset proxy, includes as many or more relevant captions than PoS, due to it considering synonyms as equivalent.
This is most evident for \epic.

\subsection{Shortcomings of \ir{} Evaluation}
\vspace{-3pt}
\begin{figure}[t]
    \centering
    \includegraphics[width=.97\linewidth]{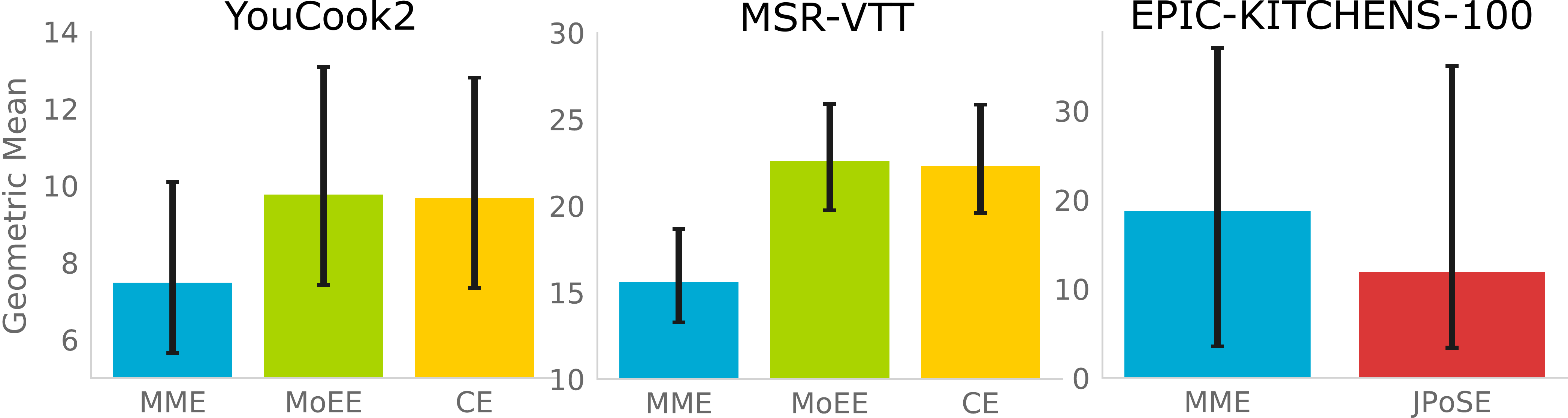}
    \caption{The min. and max. performance of baseline methods on the instance-based metric geometric mean when considering captions with $S'_{SYN}(y_i, y_j) > 0.8$ equivalent.}
    \label{fig:instance_based_error}
\end{figure}

In Sec.~\ref{sec:instance_analysis}, we analysed the shortcomings of the current approach to video retrieval that only considers a single relevant caption---the corresponding one. 
In this section, we use the semantic proxies to quantify the impact of \ir{} on the evaluation of video retrieval methods.

We consider the Geometric Mean metric, used as the prime metric in the pentathlon~\cite{albanie2020end}.
For each method, we showcase an upper/lower limit (as an error bar). 
To calculate this we consider the retrieved captions and locate the highest-retrieved caption that is above a tight threshold $S'_{SYN}(y_i, y_j) > 0.8$, per video (see Fig.~\ref{fig:sem_sim_qual_examples} for examples). 
We re-calculate the metrics, and show this as an upper limit for the method's performance.
Similarly, we locate the lowest-retrieved caption above the same threshold.
This provides the lower limit.
The figure shows the error in the evaluation metric, for each baseline on all datasets. 

From Fig.~\ref{fig:instance_based_error} we demonstrate a significant change in Geometric Mean when using the Synset-Aware proxy ($\sim$30 geometric mean for \epic, $\sim$6.0 for \msr{} and $\sim$5.0 for \yc). 
The gap between the reported performance and the upper-bound indicates that these baselines are retrieving some highly similar captions as more relevant than the ground-truth. Instance-based evaluation metrics do not account for this. 
Without considering this analysis on all relevant captions, we believe it is not possible to robustly rank methods on these benchmarks.

\begin{figure*}[t]
    \centering
    \includegraphics[width=.95\textwidth]{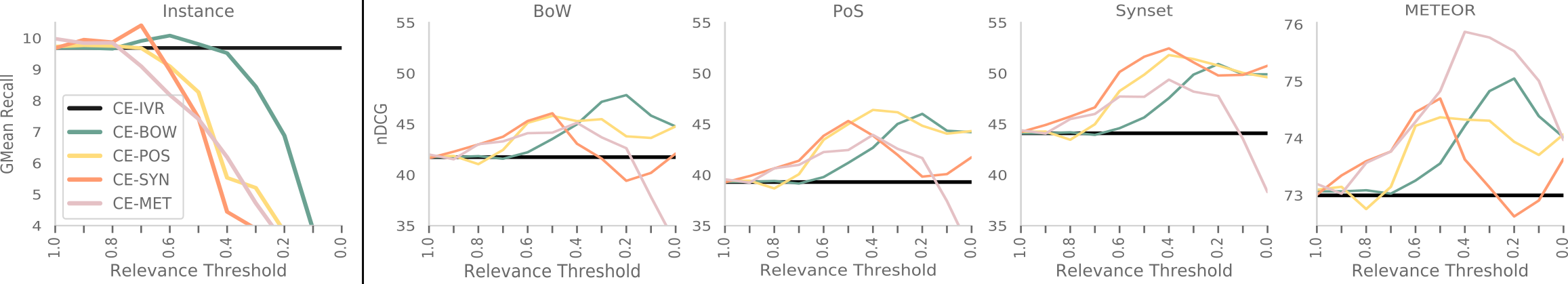}
    \caption{Training CE with semantic knowledge compared to instance-only on (left) \ir{} using Geometric Mean and (right) the four proposed semantic proxies using nDCG. Using semantic proxy in training improves performance in every case.}
    \label{fig:CE_method_ablation}
\end{figure*}

\subsection{Using Semantic Proxies for Evaluation}
\vspace{-3pt}

\begin{figure}[t]
    \centering
    \includegraphics[width=.95\linewidth]{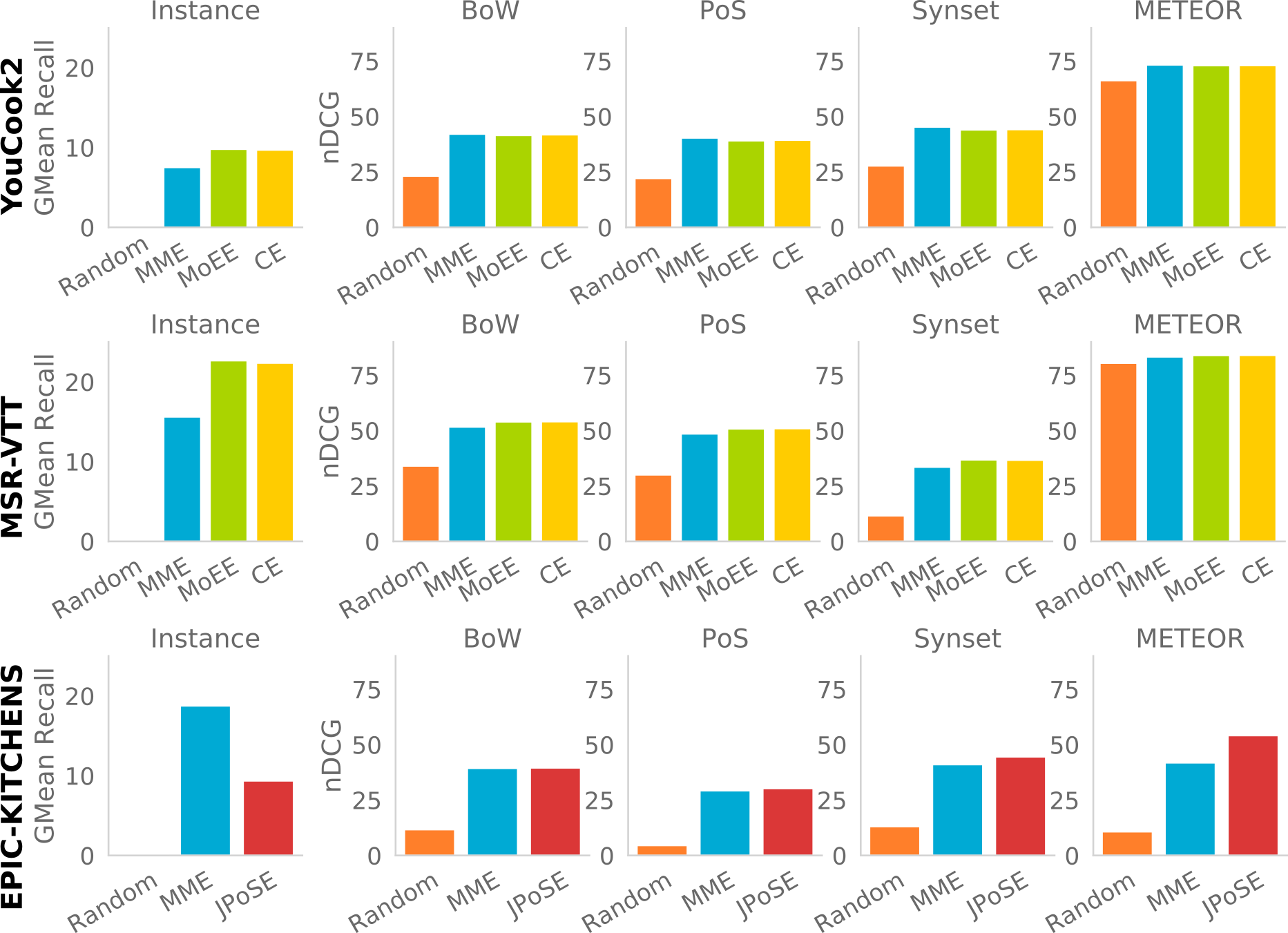}
    \caption{Evaluating the baseline methods on the proxy measures for semantic similarity (Table in supplementary).}
    \vspace{-5pt}
    \label{fig:sem_sim_results}
\end{figure}

We now evaluate \nbr{} using nDCG~(Eq.~\ref{eq:fullNDCG}) with our proposed semantic similarity proxies.
Without re-training, we evaluate nDCG on the test set, where the semantic similarities are defined using one of the four proxies in Sec.~\ref{subsec:sem_sim_proxy_def}.
We present the results in Fig.~\ref{fig:sem_sim_results} on the three datasets.

Baselines significantly outperform Random as well as the simple MME baseline on instance-based Geometric Mean. However, when semantic similarity proxies are considered, this does not hold. For almost all cases, MoEE, CE and JPoSE are comparable to MME. MME even outperforms more complex approaches (e.g. on \yc{}). 
This is critical to demonstrate, as proposed methods can produce competitive results on the problematic \ir{} setup, but may not have the same advantage in \nbr{}. 

In Fig.~\ref{fig:sem_sim_results} we can also see that the METEOR proxy leads to high nDCG values even for the Random baseline on \msr{} and \yc.
This is due to  high inter-caption similarities on average. 
Differently, JPoSE outperforms MME and Random on \epic{} for the METEOR proxy.
This suggests the hierarchy of embeddings in JPoSE improves the sentence structure matches.

While the various proxies differ in the scores they assign to captions, 
all four are suitable to showcase that tested baselines do not improve over MME. 
This demonstrates that, regardless of the semantic proxy, it is important to consider semantic similarity when assessing a method's performance, rather than using \ir{}.

\myparagraph{Choice of Semantic Proxy}
We consider all four proposed proxies to be valuable similarity metrics.
One proxy can be chosen over another for certain objectives/applications.
For example, to retrieve videos of the same recipe, BoW is useful as only videos containing the same step and ingredients are considered highly relevant. Conversely, PoS and SYN are useful when focusing on actions as they increase the importance of verbs. SYN is also particularly useful for free form captions, where synonyms are plentiful.
Multiple proxies can be considered as multiple evaluation metrics for increased robustness.

\subsection{Training with Semantic Similarity}
\vspace{-5pt}
So far, the models have been trained solely using current \ir{} losses. 
We now train with semantic knowledge using the method from Sec.~\ref{subsec:sem_sim_training}.
We limit these experiments to \yc, and the Collaborative Experts~\cite{liu2019use} (CE) baseline due to the number of models required for training for each proxy measure and threshold $T$.
We use the following labels to refer to models trained with the four proxy measures:
\textit{CE-BoW}, \textit{CE-PoS}, \textit{CE-SYN} and \textit{CE-MET} respectively. 
The original model trained using \ir, is designated as \textit{CE-\ir}.
We vary the threshold $T=\{0.1, 0.2, ..., 1\}$, showing the results in Fig.~\ref{fig:CE_method_ablation} for both \ir{} (left) and \nbr{} (right).
All plots compare to the \textit{CE-\ir} (black line).

Fig.~\ref{fig:CE_method_ablation} (left)
demonstrates that for all proxies, providing semantic information during training can increase the performance of \ir{}, however this does drop off as less similar items are treated as relevant.
As anticipated, the drop-off threshold varies per semantic proxy.

Fig.~\ref{fig:CE_method_ablation} (right) shows that
as $T$ decreases, and more captions are considered relevant in training, significant improvement in nDCG can be observed compared to \textit{CE-\ir}.
Note that the nDCG value cannot be compared across plots, as these use different semantic proxies in the evaluation. 
While the highest performance is reported when considering the same semantic proxy in both training and evaluation, training with any proxy improves results, although they peak at different thresholds.
From inspection, \textit{CE-SYN}, \textit{CE-MET} and \textit{CE-PoS} peak in performance around  $T=0.4$ whereas \textit{CE-BoW} has a peak at $T=0.2$.
When training with these specific thresholds, the models are able to best learn a semantic embedding space, which we find is agnostic of the proxy used in evaluation.

\vspace{-6pt}
\section{Conclusion}
\vspace{-4pt}
This paper highlights a critical issue in video retrieval benchmarks, which only consider instance-based (\ir) similarity between videos and captions.
We have shown experimentally and through examples the failings of the assumption used for \ir.
Instead, we propose the task of Semantic Similarity Video Retrieval (\nbr), which allows multiple captions to be relevant to a video and vice-versa, and defines non-binary similarity between items.

To avoid the infeasible burden of annotating datasets for the \nbr{} task, we propose four proxies for semantic similarity which require no additional annotation effort and scale with dataset size.
We evaluated the proxies on three datasets, using proposed evaluation and training protocols.
We have shown that incorporating semantic knowledge during training can greatly benefit model performance.
We provide a public benchmark for evaluating retrieval models on the \nbr{} task for the three datasets used in this paper at: \url{https://github.com/mwray/Semantic-Video-Retrieval}.

\vspace*{-12pt} \paragraph{Acknowledgement.} 
This work used public datasets and is supported by EPSRC UMPIRE~(EP/T004991/1).

{\small
\bibliographystyle{ieee_fullname}
\bibliography{cvprbib}
}

\appendix

\vspace*{-15pt}
\section*{Supplementary Material}
\vspace*{-5pt}

Here we provide a perceptual study that correlates the proxy measures to human annotators in Section~\ref{sec:human_study}.
Next, we provide information of the correlation between the semantic proxies in Section~\ref{sec:sem_proxy_corr}, then details on three other proxy measures showcasing their unsuitability to the three datasets in Section~\ref{sec:proxy_learnt}. Finally, we show a tabular version of Figure~\textcolor{red}{7} from the main paper for numerical comparison in future works in Section~\ref{sec:fig_7_tab}.

\vspace*{-10pt}
\section{Proxy Measures Human Agreement Study}
\vspace*{-5pt}
\label{sec:human_study}

One might wonder how do the proposed proxies in Section \textcolor{red}{4.4} correlate to human annotators. 
To answer this question, we conduct a small-scale human study.

Requesting a human to assign a score relating a video and a caption is challenging and potentially subjective, however, ranking a small number of captions for their relevance to a given video can be achieved.
We randomly select 100 videos from both the \yc{} and \msr{} datasets (we focus on these two datasets as they include the most varied captions). 
For each proposed proxy, we rank the corresponding captions by their similarity to a given video, then select the most/least relevant captions as well as the captions at the 1\textsuperscript{st}, 2\textsuperscript{nd} and 3\textsuperscript{rd} quartiles. This gives us 5 captions that are semantically distinct for the video.

We then asked 3 annotators (out of 6 total annotators) to order these 5 captions by their similarity to the given video.
We remove annotation noise by only considering consistently ordered pairs of captions---that is when all 3 annotators agree that caption A is more relevant than B.
We then report the percentage of correctly ordered pairs by the proxy, out of all consistently annotated pairs, as the `Human-to-Proxy' agreement.

Table~\ref{tab:human_study} shows the results of this human study. We note the \% of consistent pairs of captions in each case. Results demonstrate that the four proxies correlate well with human rankings, with SYN and BoW giving the best Human-to-Proxy agreement on \yc{} and \msr{} respectively.
MET has lower agreements than the other proxy measures due to it penalizing different word orders as discussed in Sec.~\textcolor{red}{5.1} of the main paper.

\begin{table}[t]
    \vspace{-10pt}
    \centering
    \resizebox{\linewidth}{!}{
    \begin{tabular}{crrrrr}
    \toprule
          & BoW & PoS & SYN & MET \\ \midrule
    \% Consistent Pairs \yc  &86.5 &78.0 &76.3 &77.3 \\ 
    \% Consistent Pairs \msr &73.1 &78.8 &75.6 &69.2 \\ \midrule
    Human Agreement \yc  &91.2&88.8&92.1&85.6\\
    Human Agreement \msr &93.7&84.8&89.7&87.5\\
    \bottomrule
    \end{tabular}}
    \caption{Human Study reporting \% of caption pairs with agreement between human and proxy on \yc{} and \msr. Note: chance is $50\%$.}
    \label{tab:human_study}
\end{table}

\vspace*{-10pt}
\section{Correlation Between Semantic Proxies}
\vspace*{-4pt}
\label{sec:sem_proxy_corr}
To determine how similar the four proposed semantic proxies are, we calculate the Pearson correlation coefficient between pairs of semantic proxies for each video in \yc{}, \msr{} and \epic{}.

\begin{figure*}[t]
    \centering
    \includegraphics[width=0.9\linewidth]{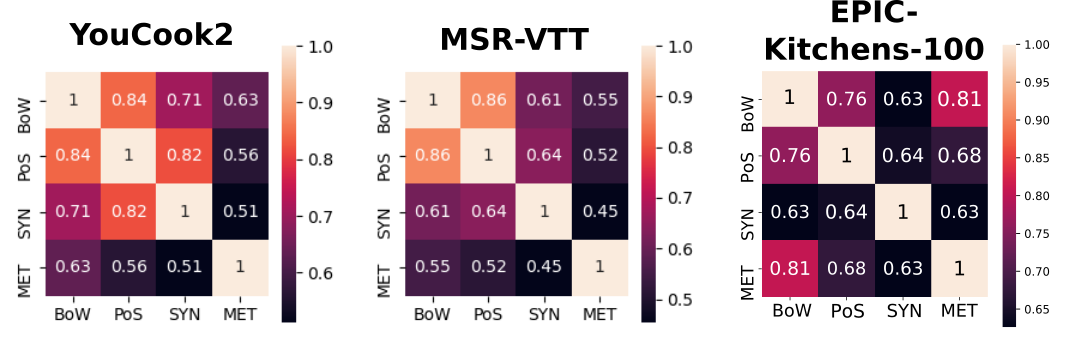}
    \caption{The average Pearson's correlation coefficient between pairs of proposed semantic proxies for \yc{}, \msr{} and \epic{}.}
    \label{fig:correlation}
\end{figure*}

Figure~\ref{fig:correlation} shows this correlation averaged over the videos within a dataset. All proposed semantic proxies have positive correlations, ranging between moderate (0.5-0.7) and high ($>$ 0.7) correlations.
We find the agreement between semantic proxies to be stronger at the lower end of the rank with the different methods consistently agreeing on which captions are irrelevant. At the higher end of the rank there tends to be some disagreements between proxies, with SYN and METEOR having the lowest correlation while BoW and PoS having the highest correlation. Importantly, the trend is consistent across the three datasets.

\begin{figure*}[t]
    \centering
    \includegraphics[width=0.96\linewidth]{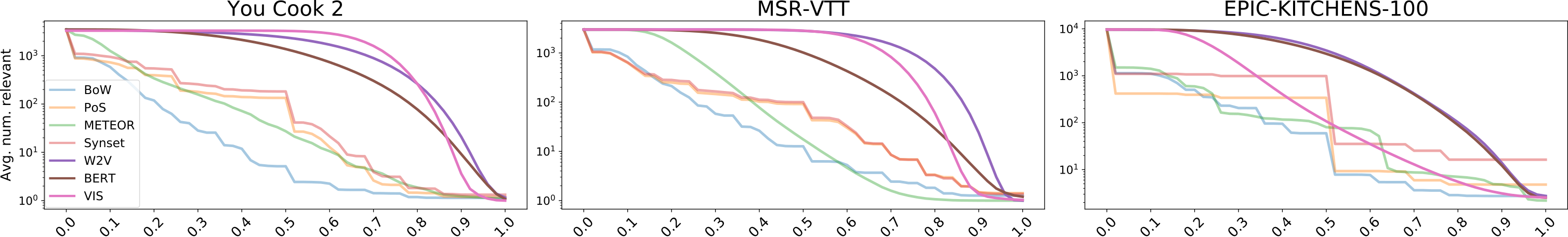}
    \caption{Average number of relevant captions for a video with a given threshold over each dataset and proxy measure including the Word2Vec (W2V), BERT and Visual (VIS).}
    \label{fig:extended_sem_histogram}
\end{figure*}

\section{Proxies from Learnt Models}
\vspace*{-5pt}
\label{sec:proxy_learnt}

\subsection{Definition}
\vspace*{-5pt}

We compare our proposed proxies (Sec \textcolor{red}{4} in the main paper) to three other proxies which use learnt features from visual or textual models.
Each proxy is defined as the cosine similarity between two vectors:
\begin{equation}
    S'(y_i, y_j) = \frac{a(y_i) \cdot a(y_j)}{||a(y_i)|| \times ||a(y_j)||}
\end{equation}
where $a(\cdot)$ is a trained model.

\vspace*{-10pt}
\paragraph{Textual Similarity}

We use two language models common in the literature to get representations: Word2Vec~\cite{mikolov2013efficient} and BERT~\cite{devlin2018bert}.
For Word2Vec, the word vectors are averaged for a sentence-level representation\footnote{We also tried using the Word Mover's Distance~\cite{kusner2015word} but achieved slightly worse results.}.
When using BERT, we extracted a sentence-level representation using the DistilBERT model from~\cite{sanh2019distilbert}.

\vspace*{-10pt}
\paragraph{Visual Similarity}

For the visual embedding proxy, we use the video features extracted from the pre-trained model. This changes Eq.~\textcolor{red}{5} in the main paper to the following:
\vspace*{-5pt}
\begin{equation}
    S_S(x_i, y_j) =
    \begin{cases}
    1 & i == j\\
    S''(x_i, x_j) &otherwise
    \end{cases}
    \label{eq:nbr}
\vspace*{-5pt}
\end{equation}
Note that a video and a caption are related here purely on the similarity between the video features, making the assumption that the visual contents of video $x_j$ offer a sufficient description of the caption $y_j$, and that the pre-trained video features offer sufficient discrimination between the videos.

\vspace*{-5pt}
\subsection{Proxy Measure Comparisons}
\vspace*{-5pt}
We show an extended version of Figure~\textcolor{red}{4} from the main paper, adding the three proxy measures from learnt models in Figure~\ref{fig:extended_sem_histogram}. We compare these for the three datasets \yc{}, \msr{} and \epic{}.

We find the average number of relevant captions per video from the three learned proxies is much higher than the proposed proxies across almost all thresholds.
With lots of captions being considered relevant, this has the effect of inflating nDCG scores.

When analysing the visual proxy, we find that the similarity is not semantic in nature. The visual proxy has high similarities between segments from the same video, further highlighting its unsuitability. Accordingly, using visual similarity from pre-trained models is not suitable as a proxy for semantic similarity.

The BERT and Word2Vec proxies similarly do not produce reasonable proxies of semantic similarities for these three datasets.
From Figure~\ref{fig:extended_sem_histogram}, both methods produce significantly more relevant captions than proposed metrics.
When analysing the results, we note that BERT and Word2Vec relate captions via their context, because of 
their training which relates words by the co-occurrence rather than their semantic relevance.
For example, `open' and `close' are often used in the same context of objects, but represent opposite actions.
Both Word2Vec and BERT would give much higher similarity to these two, despite being antonyms.

\vspace*{-5pt}
\section{Table of Figure 7}
\vspace*{-5pt}
\label{sec:fig_7_tab}

Table~\ref{tab:fig_7} shows the performance of the different baseline models on all three datasets and proxy measures.
See Section~\textcolor{red}{5.3} in the main paper for the discussion of results.

\begin{table}[hb]
    \centering
    \begin{tabular}{|c|c|r|r|r|r|r|}
    \hline
        & Proxy  & Instance & BoW & PoS & Syn & Met \\ \hline
        & Metric &  GMR & \multicolumn{4}{c|}{nDCG} \\ \hline
    \multirow{4}{*}{\rotatebox[origin=c]{90}{\yc}}
        & Random &  0.1 & 23.1 & 22.1 & 27.7 & 66.2 \\
        & MEE    &  7.5 & 42.1 & 40.3 & 45.3 & 73.3 \\
        & MoEE   &  9.8 & 41.5 & 39.1 & 44.0 & 73.0 \\
        & CE     &  9.7 & 41.8 & 39.3 & 44.1 & 73.0 \\ \hline
    \multirow{4}{*}{\rotatebox[origin=c]{90}{\msr}}
        & Random &  0.2 & 34.0 & 30.0 & 11.6 & 80.4 \\
        & MEE    & 15.7 & 51.6 & 48.5 & 33.5 & 83.3 \\
        & MoEE   & 22.7 & 53.9 & 50.8 & 36.8 & 83.9 \\
        & CE     & 22.4 & 54.0 & 50.9 & 36.7 & 84.0 \\ \hline
    \multirow{4}{*}{\rotatebox[origin=c]{90}{EPIC}}
        & Random &  0.0 & 11.7 &  4.5 & 10.7 & 13.0 \\
        & MEE    & 18.8 & 39.3 & 29.2 & 41.8 & 41.0 \\
        & JPoSE  &  9.4 & 39.5 & 30.2 & 49.0 & 44.5 \\ \hline
    \end{tabular}
    \vspace*{6pt} 
    \caption{Tabular version of Figure~\textcolor{red}{7} from the main paper. Results of evaluating the baseline methods on the different proxy measures for semantic similarity. (GMR=Geometric Mean Recall)}
    \label{tab:fig_7}
\end{table}

\end{document}